\documentclass[10pt,twocolumn,letterpaper]{article}

\usepackage{iccv}
\usepackage{times}
\usepackage{epsfig}
\usepackage{graphicx}
\usepackage{amsmath}
\usepackage{amssymb}
\usepackage{verbatim}

\usepackage{multirow}
\usepackage{booktabs}
\usepackage{caption}
\usepackage{subcaption}

\usepackage[pagebackref=true,breaklinks=true,letterpaper=true,colorlinks,bookmarks=false]{hyperref}

\iccvfinalcopy 

\def\iccvPaperID{1324} 

\ificcvfinal\fi
\begin{document}

\title{Video Frame Synthesis using Deep Voxel Flow}
\author{Ziwei Liu$^{1}$ \quad
Raymond A. Yeh$^{2}$ \quad
Xiaoou Tang$^{1}$ \quad
Yiming Liu$^{3}$\thanks{Most of the work was done when Yiming was with Google.} \quad
Aseem Agarwala$^{4}$ \\ \vspace{0.01cm}
\and 
$^{1}$The Chinese University of Hong Kong \\
{\tt\small \{lz013,xtang\}@ie.cuhk.edu.hk}
\and
$^{2}$University of Illinois at Urbana-Champaign \\
{\tt\small yeh17@illinois.edu}
\and
$^{3}$Pony.AI Inc.\\
{\tt\small yiming@pony.ai}
\and
$^{4}$Google Inc.\\
{\tt\small aseemaa@google.com}
}

\maketitle
\thispagestyle{empty}

\begin{abstract}

We address the problem of synthesizing new video frames in an existing video, either in-between existing frames (interpolation), or subsequent to them (extrapolation). This problem is challenging because video appearance and motion can be highly complex. Traditional optical-flow-based solutions often fail where flow estimation is challenging, while newer neural-network-based methods that hallucinate pixel values directly often produce blurry results. We combine the advantages of these two methods by training a deep network that learns to synthesize video frames by flowing pixel values from existing ones, which we call \textbf{deep voxel flow}. Our method requires no human supervision, and any video can be used as training data by dropping, and then learning to predict, existing frames. The technique is efficient, and can be applied at any video resolution. We demonstrate that our method produces results that both quantitatively and qualitatively improve upon the state-of-the-art.

\end{abstract}

\vspace{-8pt}
\section{Introduction}

Videos of natural scenes observe a complicated set of phenomena; objects deform and move quickly, occlude and dis-occlude each other, scene lighting changes, and cameras move. Parametric models of video appearance are often too simple to accurately model, interpolate, or extrapolate video. None the less, video interpolation, i.e., synthesizing video frames between existing ones, is a common process in video and film production. The popular commercial plug-in Twixtor\footnote{\url{http://revisionfx.com/products/twixtor/}} is used both to resample video into new frame-rates, and to produce a slow-motion effect from regular-speed video. A related problem is video extrapolation; predicting the future by synthesizing future video frames.

The traditional solution to these problems estimates optical flow between frames, and then interpolates or extrapolates along optical flow vectors. This approach is ``optical-flow-complete''; it works well when optical flow is accurate, but generates significant artifacts when it is not. A new approach~\cite{ranzato,mathieu,srivastava} uses generative convolutional neural networks (CNNs) to directly hallucinate RGB pixel values of synthesized video frames. While these techniques are promising, directly synthesizing RGB values is not yet as successful as flow-based methods, and the results are often blurry.


In this paper we aim to combine the strengths of these two approaches. Most of the pixel patches in video are near-copies of patches in nearby existing frames, and copying pixels is much easier than hallucinating them from scratch. On the other hand, an end-to-end trained deep network is an incredibly powerful tool. This is especially true for video interpolation and extrapolation, since training data is nearly infinite; any video can be used to train an unsupervised deep network.

We therefore use existing videos to train a CNN in an unsupervised fashion. We drop frames from the training videos, and employ a loss function that measures similarity between generated pixels and the ground-truth dropped frames. However, like optical-flow approaches our network generates pixels by interpolating pixel values from nearby frames. The network includes a \textit{voxel flow} layer --- a per-pixel, 3D optical flow vector across space and time in the input video. The final pixel is generated by trilinear interpolation across the input video volume (which is typically just two frames). Thus, for video interpolation, the final output pixel can be a blend of pixels from the previous and next frames. This voxel flow layer is similar to an optical flow field. However, it is only an intermediate layer, and its correctness is never directly evaluated. Thus, our method requires no optical flow supervision, which is challenging to produce at scale.

We train our method on the public UCF-101 dataset, but test it on a wide variety of videos. Our method can be applied at any resolution, since it is fully convolutional, and produces remarkably high-quality results which are significantly better than both optical flow and CNN-based methods. While our results are quantitatively better than existing methods, this improvement is especially noticeable qualitatively when viewing output videos, since existing quantitative measures are poor at measuring perceptual quality.

\section{Related Work}

Video interpolation is commonly used for video re-timing, novel-view rendering, and motion-based video compression~\cite{szeliski99, liu2014fast}. Optical flow is the most common approach to video interpolation, and frame prediction is often used to evaluate optical flow accuracy~\cite{baker2011database}. As such, the quality of flow-based interpolation depends entirely on the accuracy of flow, which is often challenged by large and fast motions. Mahajan~\etal~\cite{Mahajan:2009:MGA} explore a variation on optical flow that computes paths in the source images and copies pixel gradients along them to the interpolated images, followed by a Poisson reconstruction. Meyer~\etal~\cite{Meyer15Phase} employ a Eulerian, phase-based approach to interpolation, but the method is limited to smaller motions. 

Convolutional neural networks have been used to make recent and dramatic improvements in image and video recognition~\cite{imagenet}. They can also be used to predict optical flow~\cite{flownet}, which suggests that CNNs can understand temporal motion. However, these techniques require supervision, \ie, optical flow ground-truth. A related unsupervised approach~\cite{long} uses a CNN to predict optical flow by synthesizing interpolated frames, and then inverting the CNN. However, they do not use an optical flow layer in the network, and their end-goal is to generate optical flow. They do not numerically evaluate the interpolated frames, themselves, and qualitatively the frames appear blurry.

There are a number of papers that use CNNs to directly generate images~\cite{goodfellow} and videos~\cite{vondrick,xue2016visual}. Blur is often a problem for these generative techniques, since natural images follow a multimodal distribution, while the loss functions used often assume a Gaussian distribution. Our approach can avoid this blurring problem by copying coherent regions of pixels from existing frames. Generative CNNs can also be used to generate new views of a scene from existing photos taken at nearby viewpoints~\cite{deepstereo,deep3d}. These methods reconstruct images by separately computing depth and color layers at each hypothesized depth. This approach cannot account for scene motion, however.

Our technical approach is inspired by recent techniques for including differentiable motion layers in CNNs~\cite{jaderberg}. Optical flow layers have also been used to render novel views of objects~\cite{appearanceflow, ji2017deep} and change eye gaze direction while videoconferencing~\cite{deepwarp}. We apply this approach to video interpolation and extrapolation. LSTMs have been used to extrapolate video~\cite{srivastava}, but the results can be blurry. Mathieu~\etal~\cite{mathieu} reduce blurriness by using adversarial training~\cite{goodfellow} and unique loss functions, but the results still contain artifacts (we compare our results against this method). Finally, Finn~\etal~\cite{finn} use LSTMs and differentiable motion models to better sample the multimodal distribution of video future predictions. However, their results are still blurry, and are trained to videos in very constrained scenarios (\eg, a robot arm, or human motion within a room from a fixed camera). Our method is able to produce sharp results for widely diverse videos. Also, we do not pre-align our input videos; other video prediction papers either assume a fixed camera, or pre-align the input.

\section{Our Approach}

\begin{figure*}
  \centering
  \includegraphics[width=0.9\textwidth]{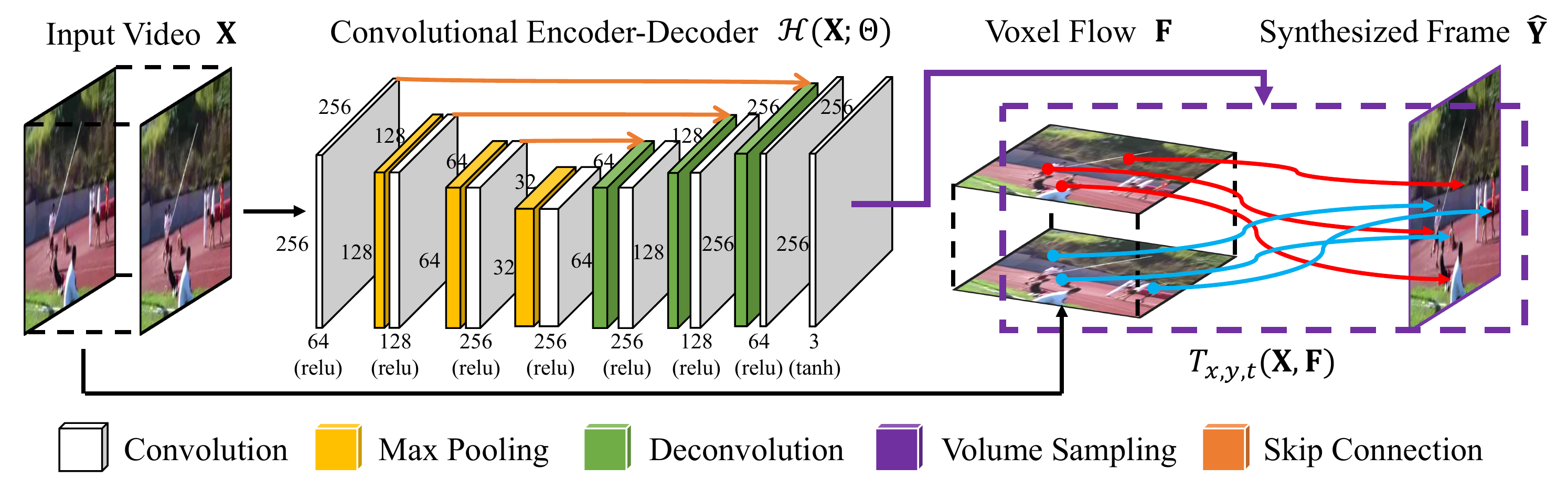}
  \caption{\small Pipeline of Deep Voxel Flow (DVF). DVF learns to synthesize a target frame from the input video. The target frame can either be in-between (interpolation) or subsequent to (extrapolation) the input video. DVF adopts a fully-convolutional encoder-decoder architecture containing three convolution layers, three deconvolution layers and one bottleneck layer. \textit{The only supervision DVF needs is the target frame which is to be synthesized.}}
  \label{fig:pipeline}
  \vspace{-8pt}
\end{figure*}

\subsection{Deep Voxel Flow}
\begin{figure*}
  \centering
  \includegraphics[width=0.95\textwidth]{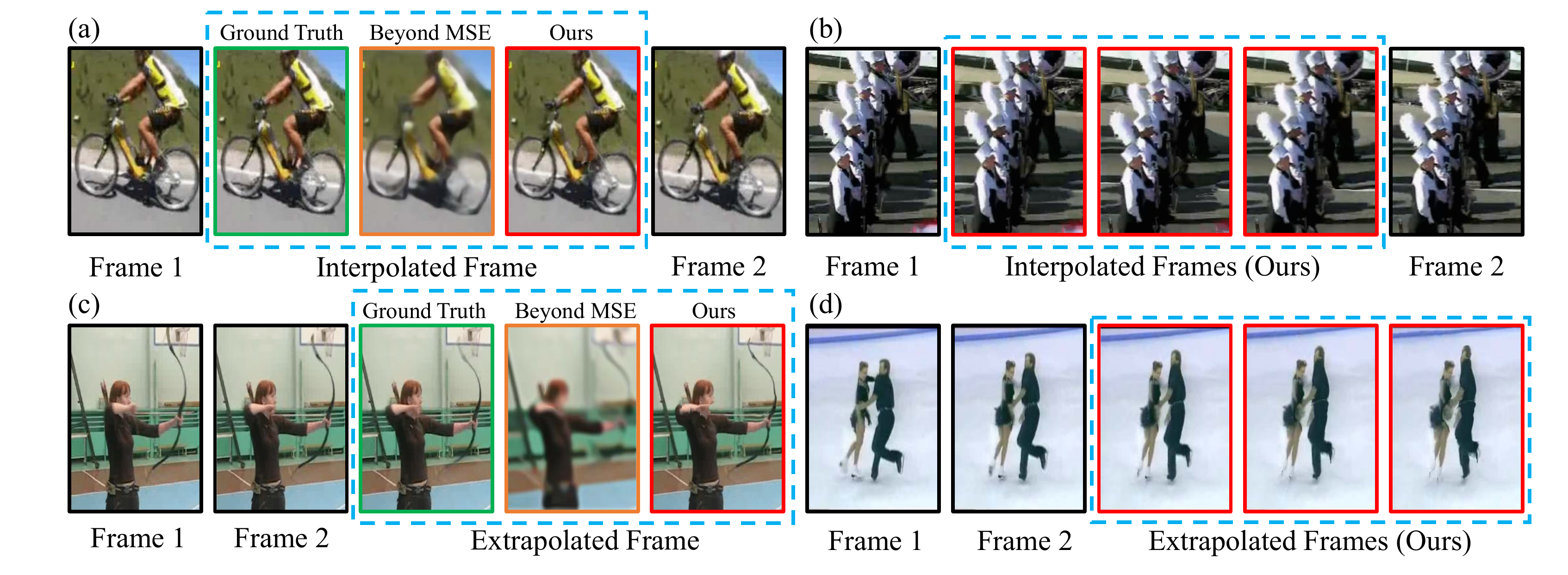}
  \caption{\small Qualitative comparisons between ground truth, Beyond MSE~\cite{mathieu} and DVF (ours) of (a) single-step video interpolation, (b) multi-step video interpolation, (c) single-step video extrapolation, (d) multi-step video extrapolation on UCF-101 dataset.}
  \label{fig:demo_ucf101}
  \vspace{-8pt}
\end{figure*}

We propose Deep Voxel Flow (DVF) --- an end-to-end fully differentiable network for video frame synthesis.
The only training data we need are triplets of consecutive video frames.
During the training process, two frames are provided as inputs and the remaining frame is used as a reconstruction target.
Our approach is self-supervised and learns to reconstruct a frame by borrowing voxels from nearby frames, which leads to more realistic and sharper results (Fig.~\ref{fig:demo_ucf101}) than techniques that hallucinate pixels from scratch.
Furthermore, due to the flexible motion modeling of our approach, no pre-processing (e.g., pre-alignment or lighting adjustment) is needed for the input videos, which is a necessary component for most existing systems~\cite{walker2016uncertain, xue2016visual}.

Fig.~\ref{fig:pipeline} illustrates the pipeline of DVF, where a convolutional encoder-decoder predicts the 3D voxel flow, and then a volume sampling layer synthesizes the desired frame, accordingly.
DVF learns to synthesize target frame $\mathbf{Y} \in \mathbb{R}^{H \times W}$ from the input video $\mathbf{X} \in \mathbb{R}^{H \times W \times L}$, where $H,~W,~L$ are the height, width and frame number of the input video.
The target frame $\mathbf{Y}$ can be the in-between frame (interpolation) or the next frame (extrapolation) of the input video. For ease of exposition we focus here on interpolation between two frames, where $L=2$.
We denote the convolutional encoder-decoder as $\mathcal{H}(\mathbf{X};\Theta)$, where $\Theta$ are the network parameters.
The output of $\mathcal{H}$ is a 3D voxel flow field $\mathbf{F}$ on a 2D grid of integer target pixel locations:
\begin{equation}
\mathbf{F} = (\Delta x, \Delta y, \Delta t) = \mathcal{H}(\mathbf{X};\Theta)~.
\end{equation}

\begin{figure}
  \centering
  \includegraphics[width=0.45\textwidth]{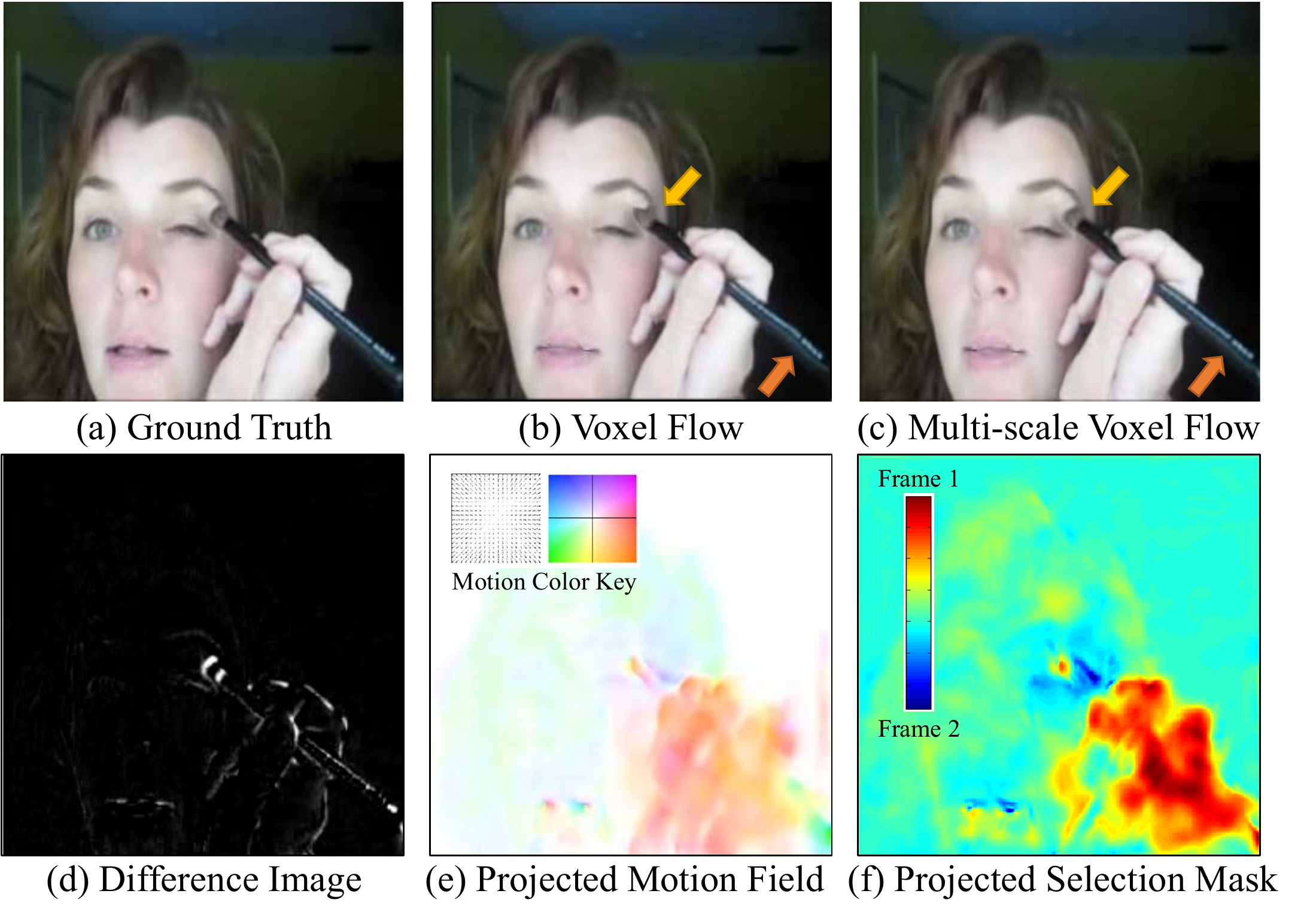}
  \caption{\small Step-by-step comparisons and visualization of DVF. (a) ground truth frame, synthesized frame by (b) voxel flow, (c) multi-scale voxel flow, (d) difference image between our result and ground truth, (e) projected motion field $\mathbf{F}_{motion}$, (f) projected selection mask $\mathbf{F}_{mask}$. Arrows draw attention to errors in the different interpolations. (\textbf{Best viewed by zooming in.})}
  \label{fig:demo_ablation}
\end{figure}

The spatial component of voxel flow $\mathbf{F}$ represents optical flow from the target frame to the next frame; the negative of this optical flow is used to identify the corresponding location in the previous frame. That is, we we assume optical flow is locally linear and temporally symmetric around the in-between frame. Specifically, we can define the absolute coordinates of the corresponding locations in the earlier and later frames as $\mathbf{L}^0 = (x - \Delta x, y - \Delta y)$ and $\mathbf{L}^1 = (x + \Delta x, y + \Delta y)$, respectively. The temporal component of voxel flow $\mathbf{F}$ is a linear blend weight between the previous and next frames to form a color in the target frame. We use this voxel flow to sample the original input video $\mathbf{X}$ with a volume sampling function $\mathcal{T}_{x,y,t}$ to form the final synthesized frame $\mathbf{\hat{Y}}$:
\begin{equation}
\mathbf{\hat{Y}} = \mathcal{T}_{x,y,t} (\mathbf{X}, \mathbf{F}) = \mathcal{T}_{x,y,t} (\mathbf{X}, \mathcal{H}(\mathbf{X};\Theta))~.
\end{equation}
The volume sampling function samples colors by interpolating within an optical-flow-aligned video volume computed from $\mathbf{X}$. Given the corresponding locations $(\mathbf{L}^0,\mathbf{L}^1)$, we construct a \textit{virtual voxel} of this volume and use trilinear interpolation from the colors at the voxel's corners to compute an output video color $\mathbf{\hat{Y}}(x, y)$. We compute the integer locations of the eight vertices of the virtual voxel in the input video $\mathbf{X}$ as:
\begin{equation}
  \begin{split}
  \mathbf{V}^{000} = &~ (\lfloor \mathbf{L}^0_x\rfloor, \lfloor \mathbf{L}^0_y\rfloor, 0) \\
  \mathbf{V}^{100} = &~ (\lceil \mathbf{L}^0_x\rceil, \lfloor \mathbf{L}^0_y\rfloor, 0) \\
  \vdots \\
  \mathbf{V}^{011} = &~ (\lfloor \mathbf{L}^1_x\rfloor, \lceil \mathbf{L}^1_y\rceil, 1) \\
  \mathbf{V}^{111} = &~ (\lceil \mathbf{L}^1_x\rceil, \lceil \mathbf{L}^1_y\rceil, 1)~,
  \end{split}
\end{equation}
where $\lfloor \cdot \rfloor$ is the floor function, and we define the temporal range for interpolation such that $t=0$ for the first input frame and $t=1$ for the second.
Given this virtual voxel, the 3D voxel flow generates each target voxel $\mathbf{\hat{Y}}(x, y)$ through trilinear interpolation:
\begin{equation}
  \mathbf{\hat{Y}}(x, y) = \mathcal{T}_{x,y,t} (\mathbf{X}, \mathbf{F}) = \sum_{i,j,k \in [0,1]} \mathbf{W}^{ijk} \mathbf{X}(\mathbf{V}^{ijk})~,
\label{eqn:trilinear}
\end{equation}
\begin{equation}
\begin{split}
  \mathbf{W}^{000} = &~ (1 - (\mathbf{L}^0_x - \lfloor \mathbf{L}^0_x \rfloor))  (1 - (\mathbf{L}^0_y - \lfloor \mathbf{L}^0_y \rfloor))  (1 - \Delta t) \\
  \mathbf{W}^{100} = &~ (\mathbf{L}^0_x - \lfloor \mathbf{L}^0_x \rfloor)  (1 - (\mathbf{L}^0_y - \lfloor \mathbf{L}^0_y \rfloor))  (1 - \Delta t) \\
  \vdots \\
  \mathbf{W}^{011} = &~ (1 - (\mathbf{L}^1_x - \lfloor \mathbf{L}^1_x \rfloor))  (\mathbf{L}^1_y - \lfloor \mathbf{L}^1_y \rfloor)  \Delta t \\
  \mathbf{W}^{111} = &~ (\mathbf{L}^1_x - \lfloor \mathbf{L}^1_x \rfloor)  (\mathbf{L}^1_y - \lfloor \mathbf{L}^1_y \rfloor)  \Delta t~,
\end{split}
\end{equation}
where $\mathbf{W}^{ijk}$ is the trilinear resampling weight.
This 3D voxel flow can be understood as the joint modeling of a 2D motion field and a mask selecting between the earlier and later frame.
Specifically, we can separate $\mathbf{F}$ into $\mathbf{F}_{motion} = (\Delta x, \Delta y)$ and $\mathbf{F}_{mask} = (\Delta t)$, as illustrated in Fig.~\ref{fig:demo_ablation} (e-f).
(These definitions are later used in Eqn.~\ref{eqn:loss} to allow different weights for spatial and temporal regularization.)

\noindent
\textbf{Network Architecture.}
DVF adopts a fully-convolutional encoder-decoder architecture, containing three convolution layers, three deconvolution layers and one bottleneck layer.
Therefore, arbitrary-sized videos can be used as inputs for DVF.
The network hyperparamters (\eg, the size of feature maps, the number of channels and activation functions) are specified in Fig.~\ref{fig:pipeline}.



For the encoder section of the network, each processing unit contains both convolution and max-pooling.
The convolution kernel sizes here are $5\times5$, $5\times5$ and $3\times3$, respectively.
The bottleneck layer is also connected by convolution with kernel size $3\times3$.
For the decoder section, each processing unit contains bilinear upsampling and convolution.
The convolution kernel sizes here are $3\times3$, $5\times5$ and $5\times5$, respectively.
To better maintain spatial information we add \emph{skip connections} between the corresponding convolution and deconvolution layers.
Specifically, the corresponding deconvolution layers and convolution layers are concatenated together before being fed forward.

\subsection{Learning}
\label{sec:learning}

For our DVF training, 
we exploit the $l_{1}$ reconstruction loss with spatial and temporal coherence regularizations to reduce visual artifacts.
Total variation (TV) regularizations are used here to enforce coherence.
Since these regularizers are imposed on the output of the network it can be easily incorporated into the back-propagation scheme.
Our overall objective function that we minimize is:
\begin{equation}
\label{eqn:loss}
\begin{split}
  \mathcal{L} = \frac{1}{N} \sum_{\langle \mathbf{X}, \mathbf{Y}\rangle \in \mathcal{D}} \big( & \|\mathbf{Y} - \mathcal{T}_{x,y,t} (\mathbf{X}, \mathbf{F})\|_{1} \\
& + \lambda_{1} \|\nabla\mathbf{F}_{motion}\|_{1} \\
& + \lambda_{2}  \|\nabla\mathbf{F}_{mask}\|_{1} \big)~,
\end{split}
\end{equation}
where $\mathcal{D}$ is the training set of all frame triplets, $N$ is its cardinality and $\mathbf{Y}$ is the target frame to be reconstructed.
$\|\nabla\mathbf{F}_{motion}\|_{1}$ is the total variation term on the $(x,y)$ components of voxel flow, and $\lambda_{1}$ is the corresponding regularization weight; similarly, $\|\nabla\mathbf{F}_{mask}\|_{1}$ is the regularizer on the temporal component of voxel flow, and $\lambda_{2}$ its weight. (We experimentally found it useful to weight the coherence of the spatial component of the flow more strongly than the temporal selection.)
To optimize the $l_{1}$ norm, we use the Charbonnier penalty function $\Phi(x) = (x^2 + \epsilon^{2})^{1/2} $ for approximation.
Here we empirically set $\lambda_{1} = 0.01$, $\lambda_{2} = 0.005$ and $\epsilon = 0.001$.


We initialize the weights in DVF using Gaussian distribution with standard deviation of $0.01$.
Learning the network is achieved via ADAM solver~\cite{kingma2014adam} with learning rate of $0.0001$, $\beta_{1} = 0.9$, $\beta_{2} = 0.999$ and batch size of $32$.
Batch normalization~\cite{ioffe2015batch} is adopted for faster convergence.

\begin{figure}
  \centering
  \includegraphics[width=0.5\textwidth]{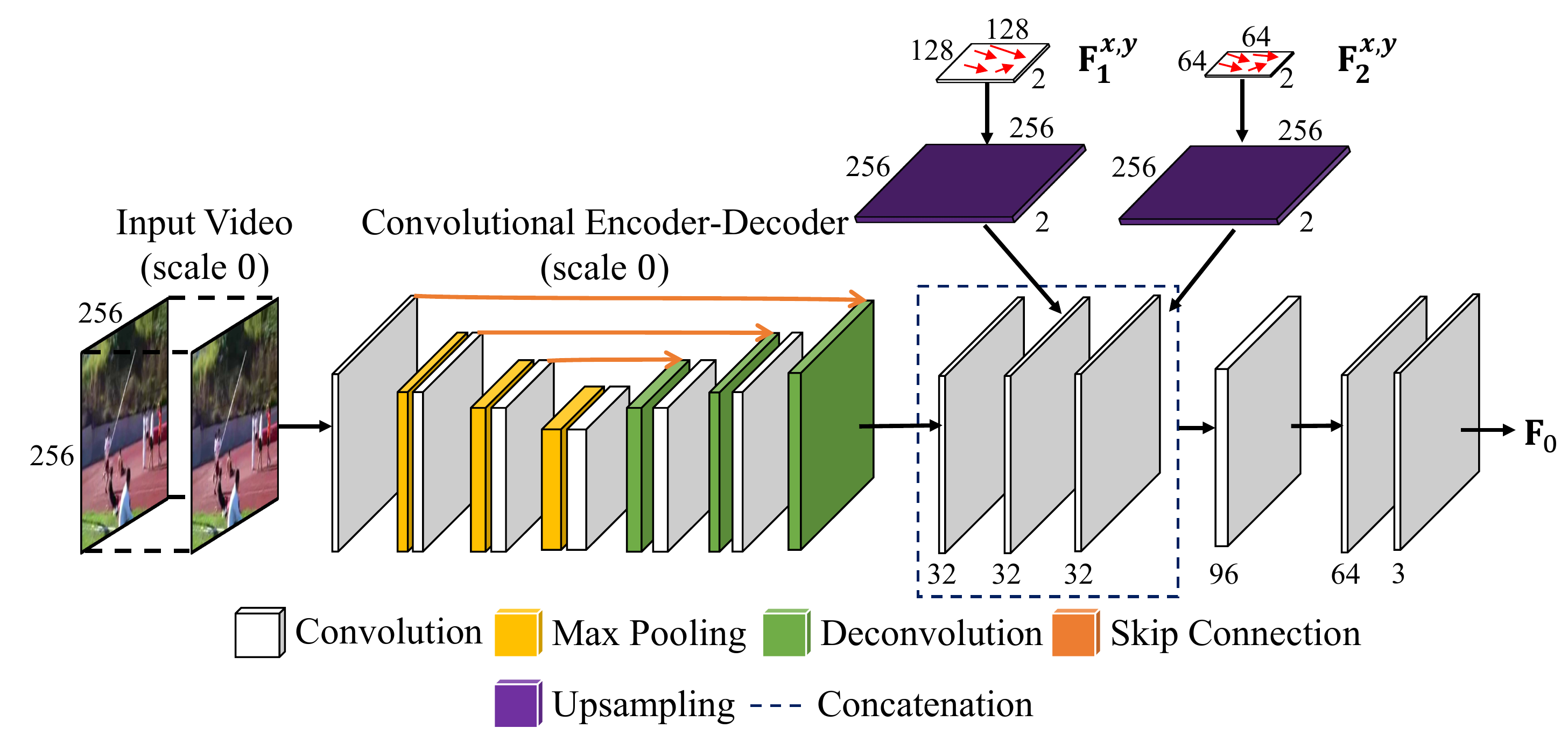}
  \caption{\small Pipeline of multi-scale Deep Voxel Flow. A series of convolutional encoder-decoder networks work on video frames from a coarse to fine scale. The spatial components of the 3D voxel flow computed at lower resolutions (here, $128 \times 128$ and $64 \times 64$) are upsampled to $256 \times 256$ and then convolved to $32$ channels. The three different resolutions are then concatenated to form a $256 \times 256 \times 96$ layer, and finally passed through several convolutional layers to form a final $256 \times 256 \times 3$ voxel flow field.}
  \vspace{-8pt}
  \label{fig:multiscale}
\end{figure}

\noindent
\textbf{Differentiable Volume Sampling.}
To make our DVF an end-to-end fully differentiable system, we define the gradients with respect to 3D voxel flow $\mathbf{F} = (\Delta x, \Delta y, \Delta t)$ so that the reconstruction error can be backpropagated through a \textit{volume sampling} layer.
Similar to~\cite{jaderberg}, the partial derivative of the synthesized voxel color $\mathbf{\hat{Y}}(x, y)$ \wrt $\Delta x$ is
\begin{equation}
\frac{\partial \mathbf{\hat{Y}}(x, y)}{\partial (\Delta x)} = \sum_{i,j,k \in [0,1]} \mathbf{E}^{ijk} \mathbf{X}(\mathbf{V}^{ijk})~,
\end{equation}
\begin{equation}
\begin{split}
  \mathbf{E}^{000} = &~ (1 - (\mathbf{L}^0_y - \lfloor \mathbf{L}^0_y \rfloor))  (1 - \Delta t) \\
  \mathbf{E}^{100} = &~ - (1 - (\mathbf{L}^0_y - \lfloor \mathbf{L}^0_y \rfloor))  (1 - \Delta t) \\
  \vdots \\
  \mathbf{E}^{011} = &~ - (\mathbf{L}^1_y - \lfloor \mathbf{L}^1_y \rfloor)  \Delta t \\
  \mathbf{E}^{111} = &~ (\mathbf{L}^1_y - \lfloor \mathbf{L}^1_y \rfloor)  \Delta t~,
\end{split}
\end{equation}
where $\mathbf{E}^{ijk}$ is the error reassignment weight \wrt $\Delta x$.
Similarly, we can compute $\partial \mathbf{\hat{Y}}(x, y) / \partial (\Delta y)$ and $\partial \mathbf{\hat{Y}}(x, y) / \partial (\Delta t)$.
This gives us a sub-differentiable sampling mechanism, allowing loss gradients to flow back to the 3D voxel flow $\mathbf{F}$.
This sampling mechanism can be implemented very efficiently by just looking at the kernel support region for each output voxel.

\subsection{Multi-scale Flow Fusion}

As stated in Sec.~\ref{sec:learning}, the gradients of reconstruction error are obtained by only looking at the kernel support region for each output voxel, which makes it hard to find large motions that fall outside the kernel.
Therefore, we propose a multi-scale Deep Voxel Flow (multi-scale DVF) to better encode both large and small motions.

Specifically, we have a series of convolutional encoder-decoder $\mathcal{H}_{N}, \mathcal{H}_{N-1}, \cdots, \mathcal{H}_{0}$ working on video frames from coarse scale $s_{N}$ to fine scale $s_{0}$, respectively.
Typically, in our experiments, we set $s_{2} = 64 \times 64$, $s_{1} = 128 \times 128$ and $s_{0} = 256 \times 256$.
In each scale $k$, the sub-network $\mathcal{H}_{k}$ predicts 3D voxel flow $\mathbf{F}_{k}$ at that resolution.
Intuitively, large motions will have a relatively small offset vector $\mathbf{F}_{k}$ in coarse scale $s_{N}$.
Thus, the sub-networks $\mathcal{H}_{N}, \cdots, \mathcal{H}_{1}$ in coarser scales $s_{N}, \cdots, s_{1}$ are capable of producing the correct multi-scale voxel flows $\mathbf{F}_{N}, \cdots, \mathbf{F}_{1}$ even for large motions.

We fuse these multi-scale voxel flows to the finest network $\mathcal{H}_{0}$ to get our final result.
The fusion is conducted by \textit{upsampling and concatenating} the multi-scale voxel flow $\mathbf{F}^{x, y}_{k}$ (only the spatial components $(\Delta x, \Delta y)$ are retained) to the final decoder layer of $\mathcal{H}_{0}$, which has the desired spatial resolution $s_{0}$.
Then, the fine-scale voxel flow $\mathbf{F}_{0}$ is obtained by further convolution on the fused flow fields.
The network architecture of multi-scale DVF is illustrated in Fig.~\ref{fig:multiscale}.
And it can be formulated as
\begin{equation}
\mathbf{\hat{Y}}_{0} = \mathcal{T} (\mathbf{X}, \mathbf{F}_{0}) = \mathcal{T} (\mathbf{X}, \mathcal{H}(\mathbf{X};\Theta, \mathbf{F}_{N}, \cdots, \mathbf{F}_{1}))~.
\end{equation}

Since each sub-network $\mathcal{H}_{k}$ is fully differentiable, the multi-scale DVF can also be trained end-to-end with reconstruction loss $\|\mathbf{Y}_{k} - \mathcal{T} (\mathbf{X}_{k}, \mathbf{F}_{k})\|_{1}$ for each scale $s_{k}$.

\subsection{Multi-step Prediction}

Our framework can be naturally extended to multi-step prediction in either interpolation or extrapolation.
For example, the goal is to predict the next $D$ frames given the current $L$ frames.
In this case, the target $\mathbf{Y}$ becomes a 3D volume ($\mathbf{Y} \in \mathbb{R}^{H \times W \times D}$) instead of a 2D frame ($\mathbf{Y} \in \mathbb{R}^{H \times W}$).
Similar to Eqn.~\ref{eqn:trilinear}, each output voxel $\mathbf{\hat{Y}}(x, y, t)$ can be obtained by performing trilinear interpolation on the input video $\mathbf{X}$, according to its projected \textit{virtual voxel}.
We have observed that spatiotemporal coherence is preserved in our output volume because convolutions across the temporal layers allow local correlations to be maintained.
%
Since multi-step prediction is more challenging, we reduce the learning rate to $0.00005$ to maintain stability when training.

\section{Experiments}

We trained Deep Voxel Flow (DVF) on videos from the UCF-101 training set~\cite{UCF101}.
We sampled frame triplets with obvious motion, creating a training set of approximately $240,000$ triplets.
Following~\cite{mathieu} and~\cite{walker2016uncertain}, both UCF-101~\cite{UCF101} and THUMOS-15~\cite{THUMOS15} test sets are used as benchmarks.
The pixel values are normalized into the range of $[-1, 1]$.
We use both PSNR and SSIM~\cite{wang2004image} (on motion regions\footnote{We use the motion masks provided by~\cite{mathieu}.}) to evaluate the image quality of video frame synthesis; higher values of PSNR and SSIM indicate better results. However, our goal is to synthesize pixels that look realistic and artifact-free to human viewers. It is well-known that existing numerical measures of visual quality are not good facsimiles of human perception, and temporal coherence cannot be evaluated in paper figures. 
We find that the visual difference in quality of our method and competing techniques is \emph{much} more significant than the numerical difference, and we include a user study in Section~\ref{sec:applications} that supports this conclusion.


\begin{table*}
\footnotesize
\parbox{.48\linewidth}{
\centering
\begin{tabular}{lcccccc}
\toprule
{\multirow{2}{*}{UCF-101~/~THUMOS-15}} & \multicolumn{2}{c}{Interpolation} & \multicolumn{2}{c}{Extrapolation} & \multicolumn{2}{c}{Extrapolation$^{\dagger}$} \\
& \multicolumn{2}{c}{\scriptsize (2 frames as input)} & \multicolumn{2}{c}{\scriptsize (2 frames as input)} & \multicolumn{2}{c}{\scriptsize (4 frames as input)} \\
\midrule
Method & PSNR & SSIM & PSNR & SSIM & PSNR & SSIM \\
\midrule
\midrule
Beyond MSE~\cite{mathieu} & 32.8~/~32.3 & 0.93~/~0.91 & 30.6~/~30.2 & 0.90~/~0.89 & 32.0 & 0.92 \\
Optical Flow & 34.2~/~33.9 & 0.95~/~0.94 & 31.3~/~31.0 & 0.92~/~0.92 & 31.6 & 0.93 \\
\midrule
Ours & \textbf{35.8}~/~\textbf{35.4} & \textbf{0.96}~/~\textbf{0.95} & \textbf{32.7}~/~\textbf{32.2} & \textbf{0.93}~/~\textbf{0.92} & \textbf{33.4} & \textbf{0.94} \\
\bottomrule
\end{tabular}
}
\hfill
\parbox{.24\linewidth}{
\centering
\begin{tabular}{lc}
\toprule
Method & $L_{1}$ Error \\
\midrule
\midrule
Recons. Views~\cite{tatarchenko2015single} & 0.492 \\
App. Flow~\cite{appearanceflow} & 0.471 \\
\midrule
Ours (w/o ft.) & 0.336 \\ 
Ours & \textbf{0.178} \\
\bottomrule
\end{tabular}
}
\caption{\small \textit{Left}: Performance (PSNR and SSIM) of frame synthesis on UCF-101 and THUMOS-15 dataset. The higher the better. ``Optical Flow'' is EpicFlow~\cite{epicflow} for all experiments except ``Extrapolation$^{\dagger}$'', whose number is taken from~\cite{mathieu}, which uses Dollar~\etal~\cite{PMT}. ``Extrapolation$^{\dagger}$'' employs the same setting as that in~\cite{mathieu}, \ie, using four frames as input and adopting the same amount of network parameters.  \textit{Right}: Performance ($L_{1}$ error) of view synthesis on KITTI dataset, with and without fine-tuning. The lower the better.}
\vspace{-6pt}
\label{tab:benchmark}
\end{table*}

\noindent
\textbf{Competing Methods.}
We compare our approach against several methods, including the state-of-the-art optical flow technique EpicFlow~\cite{epicflow}.
To synthesize the in-between images given the computed flow fields we apply the interpolation algorithm used in the Middlebury interpolation benchmark~\cite{baker2011database}.
%
%
For the CNN-based methods, we compare DVF to Beyond MSE~\cite{mathieu}, which achieved the best-performing results in video prediction. 
However, their method is trained using $4$ input frames, whereas ours uses only $2$. We therefore try both numbers of input frames. The comparisons are performed under two settings.
First, we use their best-performing loss (ADV+GDL), and replace the backbone networks in Beyond MSE~\cite{mathieu} with ours and train using the same data and protocol as in DVF (\ie, $2$ frames as input on UCF-101).
Second, we adapt DVF to their setting (\ie, using $4$ frames as input and adopting the same number of network parameters
) and directly benchmark against the numbers reported in~\cite{mathieu}.


\begin{figure}
  \centering
  \includegraphics[width=0.45\textwidth]{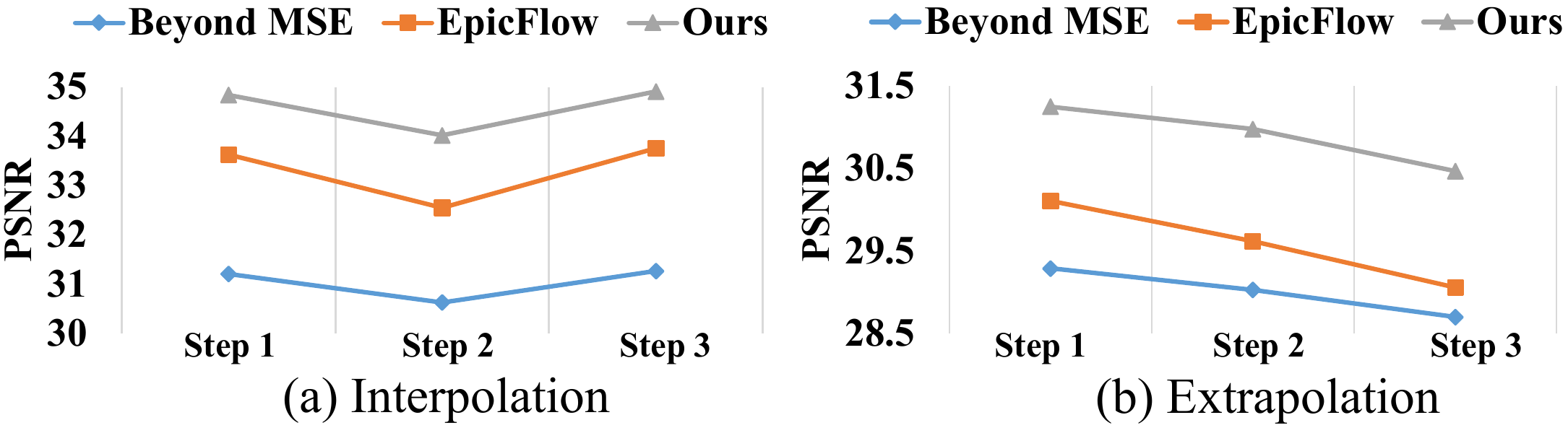}
  \vspace{-8pt}
  \caption{\small Performance comparisons on (a) multi-step interpolation and (b) multi-step extrapolation.}
  \label{fig:ablation_multistep}
\end{figure}

\begin{figure}
  \centering
  \includegraphics[width=0.45\textwidth]{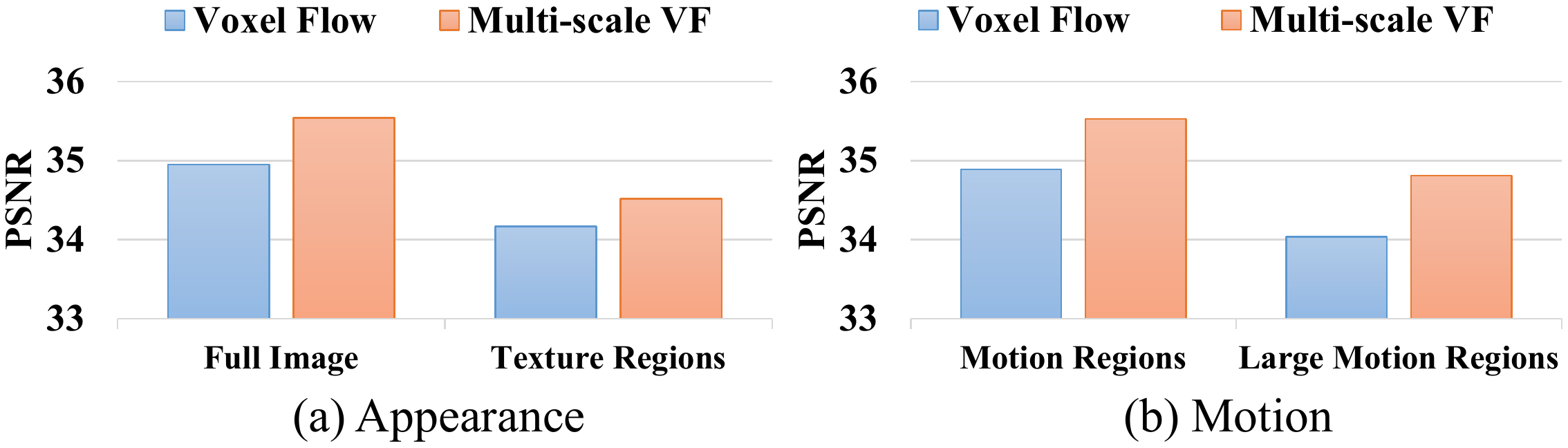}
  \vspace{-8pt}
  \caption{\small Ablation study of (a) appearance modeling and (b) motion modeling.}
  \vspace{-8pt}
  \label{fig:ablation_multiscale}
\end{figure}

\noindent
\textbf{Results.}
As shown in Table~\ref{tab:benchmark} (\textit{left}), our method outperforms the baselines for video interpolation.
Beyond MSE is a hallucination-based method and produces blurry predictions.
EpicFlow outperforms Beyond MSE by $1.4$dB because it copies pixel based on estimated flow fields. 
Our approach further improves the results by $1.6$dB.
Some qualitative comparisons are provided in Fig.~\ref{fig:demo_ucf101} (a).

Video extrapolation results are shown in Table~\ref{tab:benchmark} (\textit{middle}).
The gap between Beyond MSE and EpicFlow shrinks to $0.7$dB for video extrapolation since this task requires more semantic inference, which is a strength of deep models.
Our approach combines the advantages of both, and achieves the best performance ($32.7$dB).
Qualitative comparisons are provided in Fig.~\ref{fig:demo_ucf101} (c).

Finally, we explore the possibility of multi-step prediction, i.e., interpolate/extrapolate three frames ($step = 1, 2, 3$) at a time instead of one.
From Fig.~\ref{fig:ablation_multistep}, we can see that our approach consistently outperforms other alternatives along all time steps.
The advantages become even larger when evaluating on long-range predictions (e.g., $step = 3$ in extrapolation).
DVF is able to learn long-term temporal dependencies through large-scale unsupervised training.
The qualitative illustrations are provided in Fig.~\ref{fig:demo_ucf101} (b)(d).

\subsection{Effectiveness of Multi-scale Voxel Flow}
\label{sec:ablation}

In this section, we demonstrate the merits of Multi-scale Voxel Flow (Multi-scale VF); 
specifically, we examine results separately along two axes: appearance, and motion.
For appearance modeling, we identify the texture regions by local edge magnitude.  
For motion modeling, we identify large motion regions according to the flow maps provided by~\cite{epicflow}.
Fig.~\ref{fig:ablation_multiscale} compares the PSNR performance on UCF-101 test set without and with multi-scale voxel flow.
The multi-scale architecture further enables DVF to deal with large motions, as shown in Fig.~\ref{fig:ablation_multiscale} (b). Large motions become small after downsampling, and these motion estimates are mixed with higher-resolution estimates at the final layer of our network. The plots show that the multi-scale architecture add the most benefit in large-motion regions.

{We also validate the effectiveness of \emph{skip connections}. Intuitively, concatenating feature maps from lower layers, which have larger spatial resolution, helps the network recover more spatial details in its output. To confirm this claim, we conducted an additional ablation study, showing that removing skip connections reduced the PSNR performance by $1.1$dB.}

\subsection{Generalization to View Synthesis}

Here we demonstrate that DVF can be readily generalized to view synthesis even without re-training.
We directly apply the model trained on UCF-101 to the view synthesis task, with the caveat that we only produce half-way in-between views, whereas general view synthesis techniques can render arbitrary viewpoints.
The KITTI odometry dataset~\cite{geiger2012we} is used here for evaluation, following~\cite{appearanceflow}.

\begin{figure}
  \centering
  \includegraphics[width=0.45\textwidth]{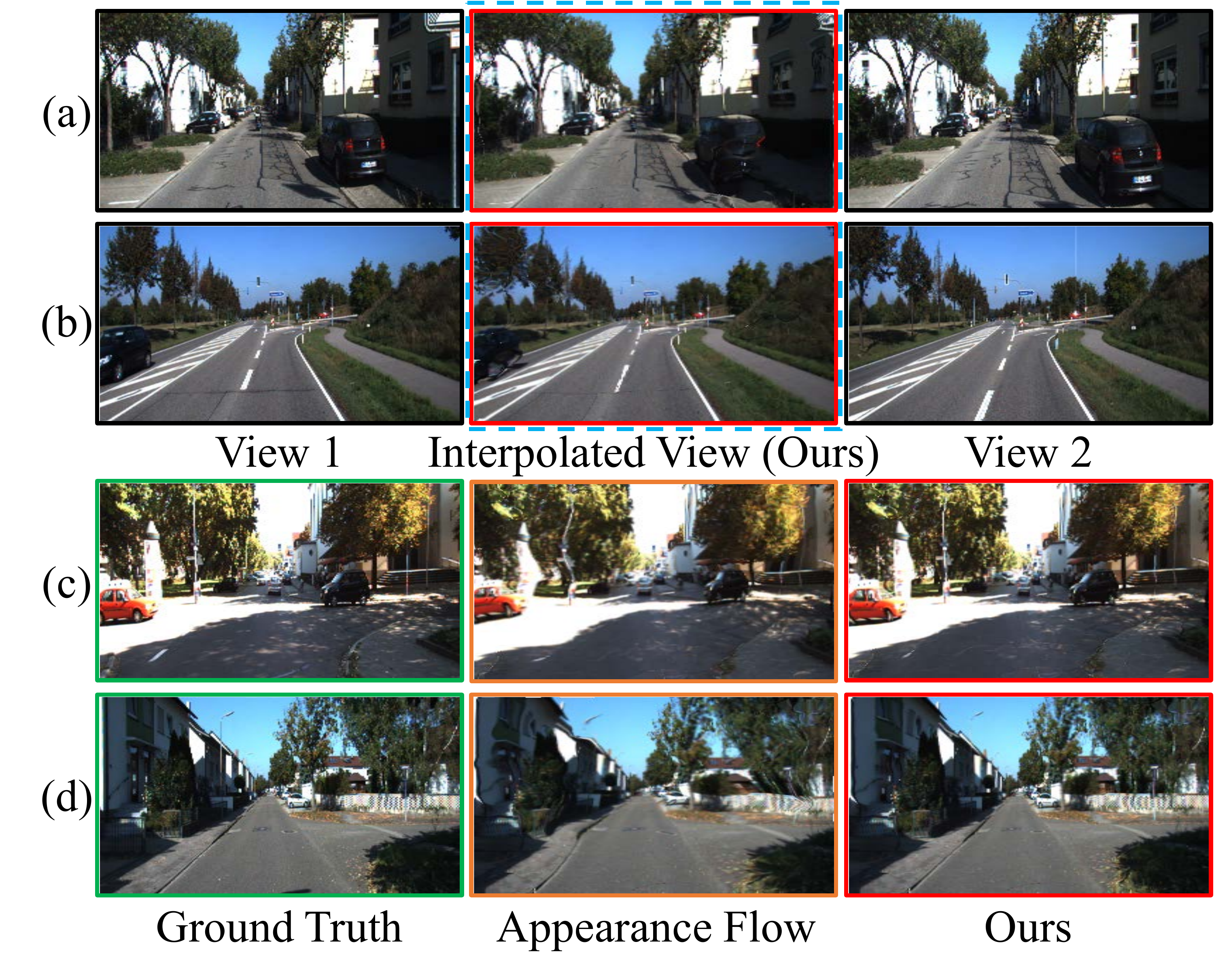}
  \caption{\small Several examples and comparisons for view synthesis on the KITTI dataset. Rows (a-b) show two examples of interpolating large viewpoint changes. Rows (c-d) compare ground truth, Appearance Flow, and our method for two other examples. Our method performs better than Appearance Flow, e.g., on the street lamp (c) and trees (d).}
  \label{fig:demo_kitti}
\end{figure}

Table~\ref{tab:benchmark} (\textit{right}) lists the performance comparisons of different methods.
Surprisingly, without fine-tuning, our approach already outperforms~\cite{tatarchenko2015single} and~\cite{appearanceflow} by $0.164$ and $0.135$ respectively.
We find that fine-tuning on the KIITI training set could further reduce the reconstruction error.
Note that KITTI dataset exhibits large camera motion, which is much different from our original training data. 
(UCF-101 mainly focuses on human actions.)
This observation implies that voxel flow has good generalization ability and can be used as a universal frame/view synthesizer.
The qualitative comparisons are provided in Fig.~\ref{fig:demo_kitti}.


\begin{table}
\footnotesize
\parbox{.48\linewidth}{
\centering
\begin{tabular}{lc}
\toprule
Method & EPE \\
\midrule
\midrule
LD Flow~\cite{brox2011large} & 12.4 \\
B. Basics~\cite{yu2016back} & 9.9 \\
FlowNet~\cite{dosovitskiy2015flownet} & 9.1 \\
EpicFlow~\cite{epicflow} & 3.8 \\
\midrule
Ours (w/o ft.) & 14.6 \\
Ours & 9.5 \\
\bottomrule
\end{tabular}
}
\hfill
\parbox{.48\linewidth}{
\centering
\begin{tabular}{lc}
\toprule
Method & Acc. \\
\midrule
\midrule
Random & 39.1 \\
Unsup. Video~\cite{wang2015unsupervised} & 43.8 \\
Shuffle\&Learn~\cite{misra2016shuffle} & 50.2 \\
ImageNet~\cite{karpathy2014large} & 63.3 \\
\midrule
Ours (w/o ft.) & 48.7 \\
Ours & 52.4 \\
\bottomrule
\end{tabular}
}
\caption{\small \textit{Left}: Endpoint error of flow estimation on KITTI dataset. The lower the better. \textit{Right}: Classification accuracy of action recognition on UCF-101 dataset, with and without fine-tuning. The higher the better. Note that our method is fully unsupervised.}
\label{tab:selfsup}
\vspace{-8pt}
\end{table}

\subsection{Frame Synthesis as Self-Supervision}

In addition to making progress on the quality of video interpolation/extrapolation, we demonstrate that video frame synthesis can serve as a self-supervision task for representation learning.
Here, the internal representation learned by DVF is applied to unsupervised flow estimation and pre-training of action recognition.

\noindent
\textbf{As Unsupervised Flow Estimation.}
Recall that 3D voxel flow can be projected into a 2D motion field, which is illustrated in Fig.~\ref{fig:demo_ablation} (e).
We quantitatively evaluate the flow estimation of DVF by comparing the projected 2D motion field to the ground truth optical flow field.
The KITTI flow 2012 dataset~\cite{geiger2012we} is used as a test set.
Table~\ref{tab:selfsup} (\textit{left}) reports the average endpoint error (EPE) over all the labeled pixels.
After fine-tuning, the unsupervised flow generated by DVF surpasses traditional methods~\cite{brox2011large} and performs comparably to some of the supervised deep models~\cite{dosovitskiy2015flownet}.
Learning to synthesize frames on a large-scale video corpus can encode essential motion information into our model.

\noindent
\textbf{As Unsupervised Representation Learning.}
%
Here we replace the reconstruction layers in DVF with classification layers (i.e., fully-connected layer + softmax loss).
The model is fine-tuned and tested with an action recognition loss on the UCF-101 dataset (split-1)~\cite{UCF101}.
This is equivalent to using frame synthesis by voxel flow as a pre-training task.
As demonstrated in Table~\ref{tab:selfsup} (\textit{right}), our approach outperforms random initialization by a large margin and also shows superior performance to other representation learning alternatives~\cite{wang2015unsupervised}.
To synthesize frames using voxel flow, DVF has to encode both appearance and motion information, which implicitly mimics a two-stream CNN~\cite{simonyan2014two}.

\begin{figure*}
  \centering
  \includegraphics[width=0.85\textwidth]{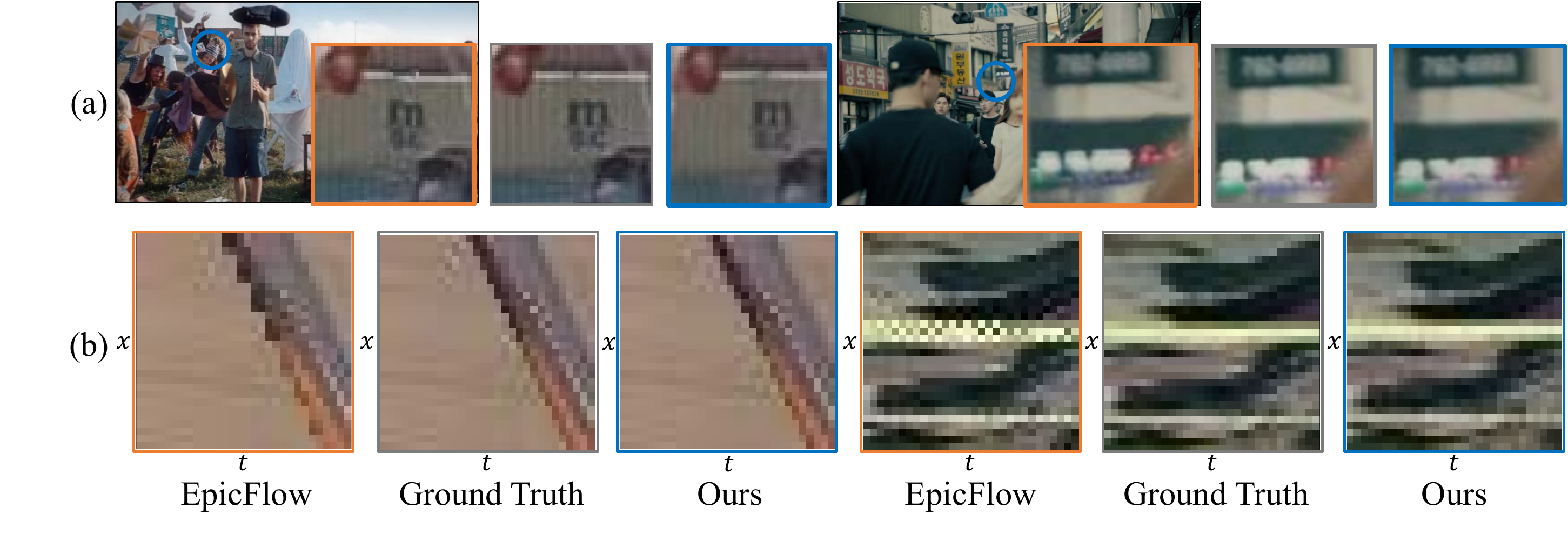}
  \vspace{-6pt}
  \caption{\small Visual quality comparisons between EpicFlow, ground truth and our approach. Row (a) shows several single frames from the output videos. Row (b) shows close-ups of $xt$ slices of each output video (rather than single frames, which are $xy$ slices). From this visualization, it can be seen that the EpicFlow output is more jagged across time. 
  }
  \vspace{-6pt}
  \label{fig:visualquality}
\end{figure*}

\begin{figure*}
  \centering
  \includegraphics[width=0.85\textwidth]{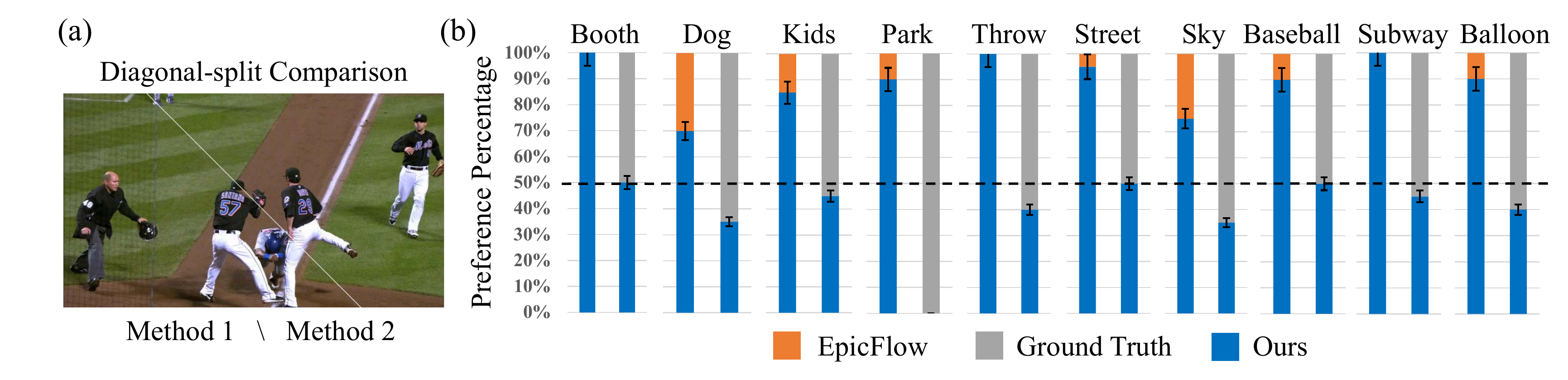}
  \vspace{-6pt}
  \caption{\small (a) Side-by-side comparison of video sequences with a diagonal-split stitch (order randomized), (b) user study results of our approach against both EpicFlow and ground truth. 95\% confidence intervals are used as error bars.}
  \vspace{-6pt}
  \label{fig:userstudy}
\end{figure*}

\subsection{Applications}
\label{sec:applications}
DVF can be used to produce slow-motion effects on high-definition (HD) videos.
We collect HD videos ($1080\times720$, $30$fps) from the web with various content and motion types as our real-world benchmark.
We drop every other frame to act as ground truth.
Note that the model used here is trained on the UCF-101 dataset without any further adaptation.
Since the DVF is fully-convolutional, it can be applied to videos of an arbitrary size.
{More video quality comparisons are available on our project page\footnote{\url{https://liuziwei7.github.io/projects/VoxelFlow}}.}

\noindent
\textbf{Visual Comparisons.}
Existing video slo-mo software relies on explicit optical flow estimation to generate in-between frames.
Thus, we choose EpicFlow~\cite{epicflow} to serve as a strong baseline.
Fig.~\ref{fig:visualquality} illustrates slo-mo effects on the ``Throw'' and ``Street'' sequences, respectively.
Both techniques tend to produce spatially coherent results, though our method performs even better.
For example, in the ``Throw'' sequence, DVF maintains the structure of the logo, while in the ``Street'' sequence, DVF can better handle the occlusion between the pedestrian and the advertisement.
However, the advantage is much more obvious when the temporal axis is examined. We show this advantage in static form by showing $xt$ slices of the interpolated videos (Fig.~\ref{fig:visualquality} (c)); the EpicFlow results are much more jagged across time. Our observation is that EpicFlow often produces zero-length flow vectors for confusing motions, leading to spatial coherence but temporal discontinuities. Deep learning is, in general, able to produce more temporally smooth results than linearly scaling optical flow vectors.

\noindent
\textbf{User Study.}
We conducted a user study on the final slo-mo video sequences to objectively compare the quality of different methods.
We compare DVF against both EpicFlow and ground truth.
For side-by-side comparisons, synthesized videos of the two different methods are stitched together using a diagonal split, as illustrated in Fig.~\ref{fig:userstudy} (a).
The left/right positions are randomly placed.
Twenty subjects were enrolled in this user study; they had no previous experience with computer vision.
We asked participants to select their preferences on $10$ stitched video sequences, i.e., to determine whether the left-side or right-side videos were more visually pleasant.
As Fig.~\ref{fig:userstudy} (b) shows, our approach is significantly preferred to EpicFlow among all testing sequences. 
{For the null hypothesis: ``there is no difference between EpicFlow results and our results'', 
the p-value is $p < 0.00001$, and the hypothesis can be safely rejected.}
Moreover, for half of the sequences participants choose the result of our method roughly equally as often as the ground truth, which suggests that they are of equal visual quality. 
{For the null hypothesis: ``there is no difference between our results and ground truth'', 
the p-value is $0.838193$; statistical significance is not reached to safely reject the null hypothesis in this case.}
Overall, we conclude that DVF is capable of generating high-quality slo-mo effects across a wide range of videos.

\noindent
\textbf{Failure Cases.} 
The most typical failure mode of DVF is in scenes with repetitive patterns (e.g., the ``Park'' sequence).
In these cases, it is ambiguous to determine the true source voxel to copy by just referring to RGB differences. 
Stronger regularization terms can be added to address this problem.



\section{Discussion}

In this paper, we propose an end-to-end deep network, Deep Voxel Flow (DVF), for video frame synthesis. Our method is able to copy pixels from existing video frames, rather than hallucinate them from scratch. On the other hand, our method can be trained in an unsupervised manner using any video. Our experiments show that this approach improves upon both optical flow and recent CNN techniques for interpolating and extrapolating video.
In the future, it may useful to combine flow layers with pure synthesis layers to better predict pixels that cannot be copied from other video frames. Also, the way we extend our method to multi-frame prediction is fairly simple; there are a number of interesting alternatives, such as using the desired temporal step (e.g., $t=.25$ for the first out of three interpolated frames) as an input to the network. Compressing our network so that it may be run on a mobile device is also a direction we hope to explore.

{\small
\bibliographystyle{ieee}
\bibliography{egbib}
}

\end{document}